\documentclass[journal]{IEEEtran}
\usepackage{framed,multirow}
\usepackage{booktabs}
\usepackage{amssymb}
\usepackage{latexsym}

\usepackage{url}
\usepackage{xcolor}
\definecolor{newcolor}{rgb}{.8,.349,.1}

\usepackage{subfigure}
\usepackage{enumerate}
\usepackage{pbox}
\usepackage{graphicx}
\usepackage{hyperref}
\usepackage{array}
\newcolumntype{P}[1]{>{\centering\arraybackslash}p{#1}}

\begin{document}
\title{Multi-Channel CNN-based Object Detection for Enhanced Situation Awareness}

\author{
	\vskip 1em
	{
	Shuo Liu, \emph{Student Member}, \emph{IEEE},
	Zheng Liu, \emph{Senior Member}, \emph{IEEE}
	}

\thanks{S. Liu and Z. Liu (Corresponding author) are with Department of School of Engineering | Faculty of Applied Science, University of British Columbia Okanagan Campus, Kelowna, Canada (email: zheng.liu@ubc.ca).}
	
}

\maketitle
	
\begin{abstract} 
Object Detection is critical for automatic military operations. However, the performance of current object detection algorithms is deficient in terms of the requirements in military scenarios. This is mainly because the object presence is hard to detect due to the indistinguishable appearance and dramatic changes of object's size which is determined by the distance to the detection sensors.   Recent advances in deep learning have achieved promising results in many challenging tasks. The state-of-the-art in object detection is represented by convolutional neural networks (CNNs), such as the fast R-CNN algorithm. These CNN-based methods improve the detection performance significantly on several public generic object detection datasets. However, their performance on detecting small objects or undistinguishable objects in visible spectrum images is still insufficient. In this study, we propose a novel detection algorithm for military objects by fusing multi-channel CNNs. We combine spatial, temporal and thermal information by generating a three-channel image, and they will be fused as CNN feature maps in an unsupervised manner. The backbone of our object detection framework is from the fast R-CNN algorithm, and we utilize cross-domain transfer learning technique to fine-tune the CNN model on generated multi-channel images. In the experiments, we validated the proposed method with the images from SENSIAC (Military Sensing Information Analysis Centre) database and compared it with the state-of-the-art. The experimental results demonstrated the effectiveness of the proposed method on both accuracy and computational efficiency.
\end{abstract}


\markboth{}%
{}

\definecolor{limegreen}{rgb}{0.2, 0.8, 0.2}
\definecolor{forestgreen}{rgb}{0.13, 0.55, 0.13}
\definecolor{greenhtml}{rgb}{0.0, 0.5, 0.0}

\section{Introduction}
Automatic target detection (ATD) is a key technology for automatic military operations and surveillance missions. In military mission, sensors can be placed on the ground, mounted on unmanned aerial vehicles(UAVs) and unmanned ground vehicles (UGVs) to acquire sensory data. Then the data will be processed using ATD algorithms which aim to find bounding boxes where the targets may be located. Fast and accurate object detector can increase lethality and survivability of the weapons platform/soldier. Whether the tactical scenario is the onslaught of an array of combat vehicles coming through the Fulda Gap, which was feared during the Cold War \cite{Ratches2011ReviewTasks}, or the identification of humans with intent to kill in an urban scene, the identification of the threat for avoidance and engagement is paramount to survival and threat neutralization.

\begin{figure}
\centering
\includegraphics[width=\linewidth]{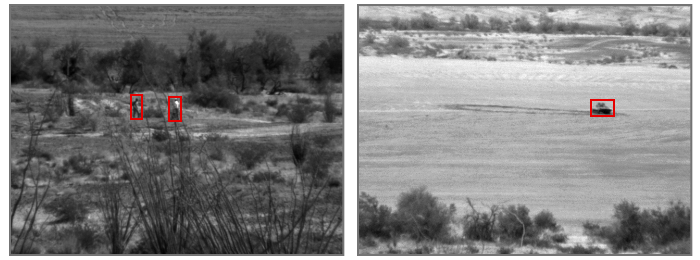}
\caption{\textbf{Left}: the appearance of target is undistinguishable from background environment. \textbf{Right}: the scale of target is various dramatically. }
\label{image-sample}
\end{figure}

Numerous ATD algorithms have been proposed during past decades. Generally, these algorithms can be classified into two main categories: 1) background modeling approaches, 2) and learning-based approaches. 

Background modeling approaches assume that background pixels have a similar color (or intensity) over time in a fixed camera, and the background model is built on this assumption. The background is abstracted from the input image, and the foreground (moving objects) region is determined by marking the pixels in which a significant difference occurs. In\cite{Sheikh2005BayesianScenes}, the authors modeled the background using a Kernel density estimation (KDE) method over a joint domain-range representation of image pixels. Multilayer codebook-based background subtraction (MCBS) model was proposed in\cite{Guo2013FastDetection}, which can remove most of the non-stationary background and significantly increase the processing efficiency. Reference\cite{Chen2015ProbabilisticSystems} proposed a motion detection model based on probabilistic neural networks. Above methods are designated for the stationary camera scenario. In the works of\cite{Yun2017SceneCamera}\cite{Hu2015MovingCamera}\cite{Francisco2015ForegroundApproximation}, the authors proposed several schemes that can handle the problems in the moving camera scenario. The background modeling based methods are effective for detecting moving objects, whereas when the objects are still or moving slowly, those methods will always be unsatisfying.

Another popular category is the learning-based approaches. Traditionally, hand-engineered features like SIFT \cite{Keypoints2004DistinctiveFrom} or HOG \cite{Dalal2010ToDetection} are firstly extracted and then fed into a classifier, such as boosting\cite{P.Viola2001RapidFeatures}, support vector machine (SVM)\cite{Suykens1999LeastClassifiers} and random forest \cite{Breiman2001Randomforest2001}. The typical work in this paradigm is the deformable part models (DPM) \cite{Felzenszwalb2009ObjectModels}. More recently, Convolutional Neural Networks (CNN) raise a significant impact on ATD research community, which helped achieve promising results in many difficult object detection challenges \cite{RussakovskyImageNetChallenge} \cite{Everingham2010TheChallenge} \cite{LinMicrosoftContext}. \textit{Overfeat} \cite{Sermanet2014OverFeat:Networks} firstly utilized CNN models in a sliding window fashion on ATD task successfully. Where it has two CNNs, one for classifying if a window contains an object and the other for predict the bounding box of the object. After that, the most popular CNN-based ATD framework appeared, R-CNN \cite{GirshickRichSegmentation}, which uses a pre-trained CNN to extract features from box proposals generated by selective search \cite{Uijlings2013SelectiveRecognition}, and then it classifies them with class specific linear SVMs. The significant advantage of this work is derived from replacing hand-engineered features by CNN feature extractor. Then, the variants of R-CNN were proposed to mainly solve the problem with computational burden \cite{Girshick2016} \cite{He2015} \cite{RenFasterNetworks}.

Nevertheless, above mentioned ATD methods are only applicable to the general natural scene, and many challenges come up from the military scenario. First, the environment of battlefields is extremely complex. As shown in Figure \ref{image-sample}, the appearance of the object includes color and texture is similar to the background in left example, because soldiers always attempt to decorate themselves or their vehicles similar to the environment in order to be invisible. And due to the vast battlefield, the scale of objects always dramatically changes with their distance to sensors. Thus, those environmental factors will always limit the ability of generic object detection algorithm. Second, military ATD application always runs on the embedded platform whose computational and memory resources are limited. In this case, the ability to run at high frame rates with relatively high accuracy becomes a crucial issue for military ATD.

Several image fusion based methods were proposed to enhance target representation in literature \cite{Bhatnagar2015ASurveillance} \cite{Niu2012AirborneTransform} \cite{Han2007FusionDetection} \cite{Smeelen2014Semi-hiddenImages}. Multiple images acquired with different range of electromagnetic spectrum were fused into one image by pixel-level image fusion algorithms such as PCA-based weighted fusion \cite{Smeelen2014Semi-hiddenImages}and Discrete Wavelet Transform (DWT) \cite{Niu2012AirborneTransform}, and then fed into an ATD system. When the fused images are used in ATD tasks, there are still deficient. To address the serious limitation, we propose a novel image fusion approach to improving detector performance in the military scenario, which exploits the significant advantage of the unsupervised feature learning characteristic of CNNs. Compared with high-level image fusion, the proposed method can achieve a higher accuracy and computational efficiency. In addition, we adopted the state-of-the-art generic object detection framework into the military scenario and used a cross-domain transfer learning techniques to cover the shortage of insufficient data. In this way, the proposed framework achieved promising results on SENSIAC military dataset. 
To sum up, our contributions in this work are as follows:
\begin{enumerate}
\item Both spatial-temporal and multiple spectral bands information are employed into ATD task.
\item An unsupervised learning-based image fusion approach is proposed, which can automatically learn how to fuse the essential information from the different type of images into a set of discriminative feature maps. 
\item The proposed framework has the capability to transfer learning from visible images to our created complex fused images. Moreover, the proposed method can achieve $98.34\%$ average precision and $98.90\%$ top1 accuracy on SENSIAC military dataset.
\end{enumerate}

\section{Methodology}
\begin{figure*}[htbp]
\includegraphics[width=\textwidth]{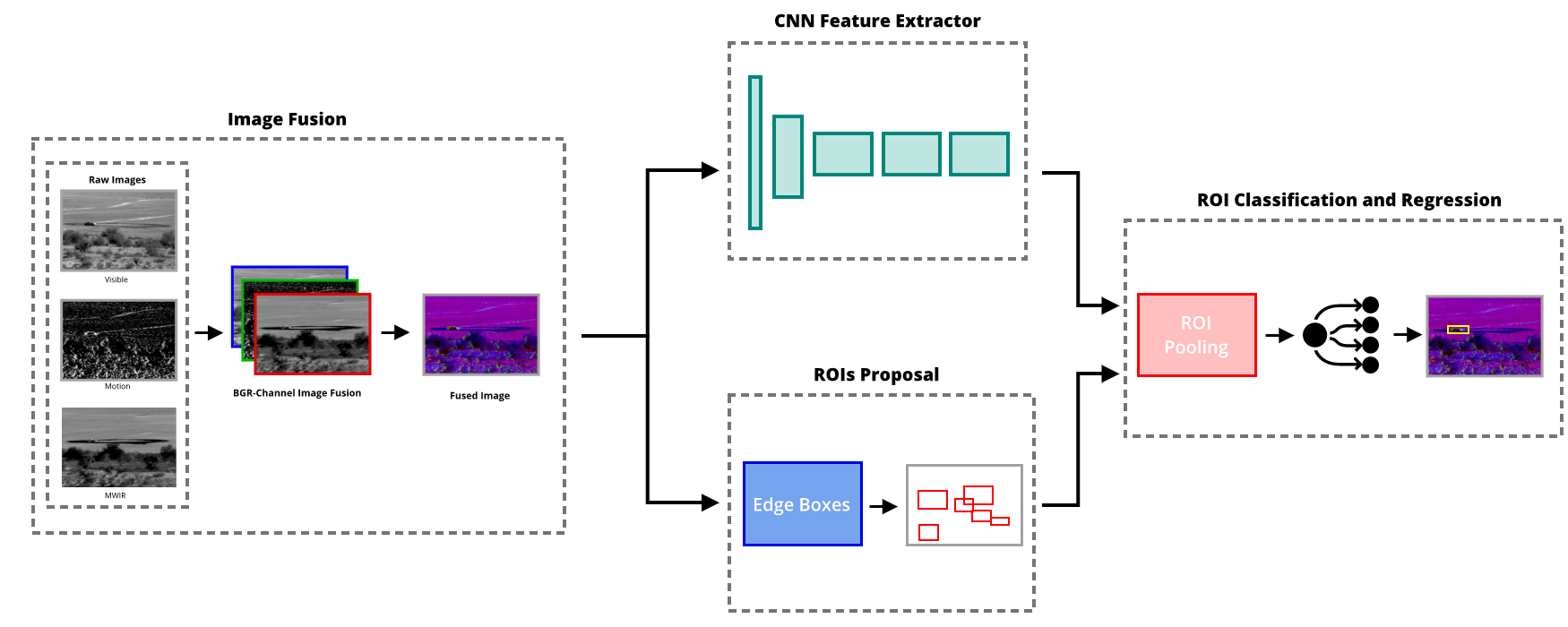}
\caption{The pipeline of proposed object detector framework, which include four main components:1) image fusion, 2) ROI proposal, 3) CNN feature extractor, and 4) ROI classification and regression. }
\centering
\label{pipeline}
\end{figure*}

In this paper, we proposed a framework, namely, image fused object detector (IFOD). As illustrated in Figure \ref{pipeline}, the whole system is composed of four modules: 1) an image fusion module, which can fuse three different type of images into a BGR image; 2) a CNN feature extractor,  used for extracting high-level semantic representations from the fused image; 3) a region of interest (ROI) proposal module manipulated on fused image is utilized for generating hundreds or thousands of candidate bounding boxes, for each ROI on feature map produced by feature extractor module; and 4) an ROI classification and regression is performed to obtain fine bounding boxes and corresponding class.

\subsection{Image Fusion}

\subsubsection{Image Selection}
Multi-sensor data often provide complementary information for context enhancement, which may further enhance the performance of object detection. In our work, we investigated two type of images from different sensors, mid-wave infrared image (MWIR) and visible image (VI), respectively. In addition to the images acquired from these two sensors, we also incorporate motion image generated from two consecutive visible frames in order to complement sufficient description of objects.

\textit{\textbf{MWIR}}: Depending on the different range of electromagnetic spectrum, the infrared (IR) spectrum can be divided into different spectral bands. Basically, the IR bands include the active IR band and the passive (thermal) IR band. The main difference between active and passive infrared bands is that the passive IR image can be acquired without any external light source whereas the active IR required that. The passive (thermal) IR band is further divided into the mid-wave infrared (3-5 \textit{um}) and the long-wave infrared (7-14 \textit{um}).In general, the mid-wave infrared (MWIR) cameras can sense temperature variations over targets and background at a long distance, and produce thermograms in the form of 2D images. Its ability to present large contrasts between cool and hot surfaces is extremely useful for many computer vision tasks such as image segmentation and object detection. However, the MWIR sensor is not sensitive to cool background. And due to low-resolution sensor arrays and the possible absence of auto-focus lens capabilities, high-frequency content of the objects like edges and texture are mostly missed.

\textit{\textbf{VI}}: The range of the electromagnetic spectrum of visible image is from 380 \textit{nm} to 750 \textit{nm}. This type of image can be easily and conveniently acquired via various kinds of general cameras. In comparison with MWIR image, the VI image is sensitive to illumination changes, preserve high-frequency information and can provide a relatively clear perspective of the environment. In most of the computer vision topics, the VI image has become major studying object for many decades. Thus, there are a large number of public VI datasets across many research areas. On the other hand, the significant drawbacks of VI image are that it has poor quality in the harsh environmental conditions with unfavourable lighting and pronounced shadows, and there is no dramatic contrast between background and foreground when the environment is extremely complicated such as the battlefield.

\begin{figure}[htbp]
\includegraphics[width=\linewidth]{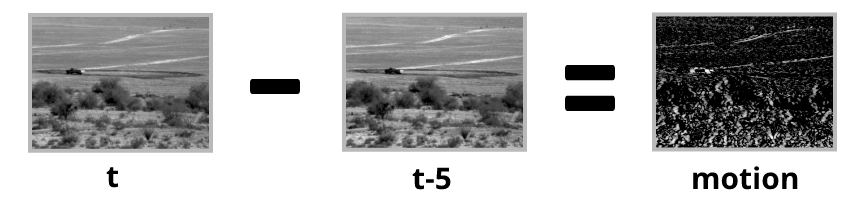}
\caption{The procedure of motion estimation: where t is the current frame and t-5 is the previous 5th frame, and the motion is what our algorithm estimate. }
\centering
\label{motion}
\end{figure}

\textit{\textbf{Motion image}}: In general, the moving objects are the targets in the battle fields. Therefore, estimating the motion of objects can provide significant cues to segment those targets.  Various motion estimation algorithms have been proposed in recent decades, such as dense optical flow methods, points correspondence methods, and background subtraction. And each of them has shown effectiveness on many computer vision tasks. However, considering the trade-off between accuracy and computational complexity, we do not opt for any of the complicated motion estimation approaches but utilize a straightforward and easier to be implemented method. The method is illustrated in Fig \ref{motion}, we estimate the motion map based on two consecutive frames. To be specific, the objective images are sampled at every 5 frames, and then force the current frame to subtract the last frame, the resulting image is the desired motion image. The method can be formulated as follow:
\begin{equation}
M_n(x,y) = | I_n(x,y) - I_{n-5}(x,y)|
\end{equation}
where $M_n(x,y)$ represents the motion value of frame $n$ at pixel point $(x,y)$ and $I_n(x,y)$ denotes the pixel value of frame $n$ at pixel point (x,y).

In this way, we do not need to consider multiple frames to estimate background, like the background subtraction methods, and only the subtraction operator is employed in this procedure, which is more efficient that other state-of-the-art methods. Even though this method can bring lots of noise in the motion image, this image still can provide enough complementary information in the later fusion stage.

\subsubsection{Fusion Methodology}

\begin{figure*}[ht]
\includegraphics[width=0.8\textwidth]{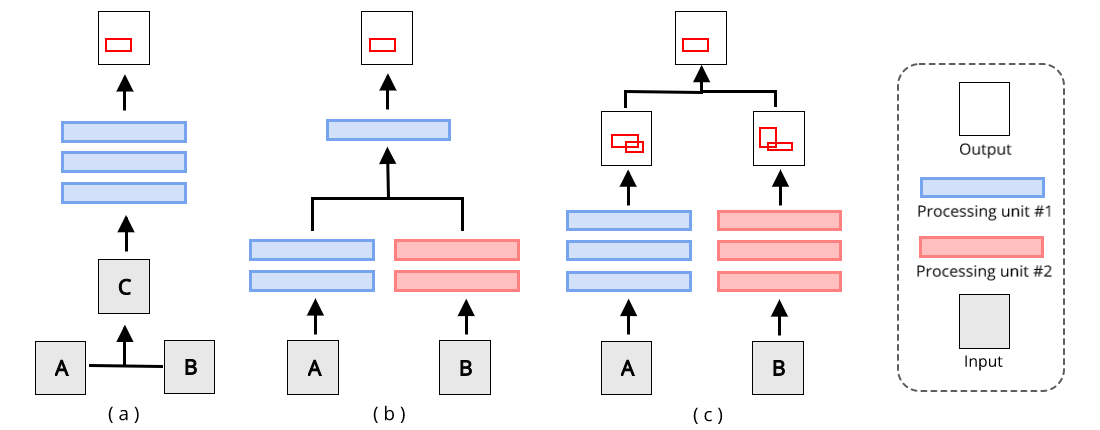}
 \centering
 \caption{Illustration of different image fusion architectures: (a) pixel-level fusion architecture; (b) feature-level fusion architecture; (c) decision-level fusion architecture. }
 \label{fusion_configurations}
\end{figure*}

Here, we formalize the possible configurations of information fusion for object detection into three categories, namely, \textit{pixel-}, \textit{feature-} and \textit{decision-} fusion architecture. An illustration is shown in Figure \ref{fusion_configurations}. Having these possibilities in mind will help to highlight the important benefits of our proposed fusion method in terms of efficiency and effectiveness.

\textbf{\textit{Pixel-level fusion}}: A typical pixel-level fusion architecture is illustrated in Figure \ref{fusion_configurations}(a). This configuration of image fusion is the lowest level techniques dealing with the pixels obtained from the sensor directly and tries to improve the visual enhancement. Typically, multiple images from different sources are combined into one single image in pixel-wise manner, after which it is fed into the object detection system to generate final results. One of the main advantages of the pixel-level fusion is their low computational complexity and easy implementation.

\textbf{\textit{Feature-level fusion}}: As an higher level fusion system to Figure \ref{fusion_configurations}(a), one might pursue Figure \ref{fusion_configurations}(b), in which different type of images are simultaneously fed into their independent lower part of the entire object detection system, which is typically called feature extractor. For instance, this lower-level system might be the hand-engineered feature extractor for the traditional object detection system, and high-level convolution layer for CNN-based system. After which the concatenated features produced by the various independent lower system are fed into one upper (decision-making) system to produce the final results. Although this feature-level fusion is usually robust to noise and misregistration, it always require almost double memory and computing resource to deal with feature fusion procedure in a parallel fashion, especially for the CNN-based methods. 

\textbf{\textit{Decision-level fusion}}: The decision-level fusion scheme illustrated in Figure \ref{fusion_configurations}(c) operates on the highest level, and refers to fusing discriminate results from different systems designed for various type images. Note that for an object detection system which usually based on machine learning algorithms, this high-level fusion probably could not establish well relationship of interior characteristics between different type of images.In addition, this method might also be practically challenging to implement as this duplication would multiply the amount of resources and running time.

In our framework, we proposed a novel image fusion approach which is similar to pixel-level fusion style, we combine multiple images into one single image while we do not conduct fusion algorithm in pixel-level stage. As you can see in the image fusion module in Figure \ref{pipeline}, firstly, the three type of raw images (MWIR, VI and Motion image) are concatenated into a BGR-style three-channel image where MWIR in red channel, motion image in green channel and VI in blue channel. Then we obtain this three-channel fused image, and it is worthy noted that we do not modify any pixel values of the raw images but just put them into independent channels of final fused image. After the fused image is obtained, we fed it into convolutional neural network (CNN) to training our object detector in end-to-end manner, meanwhile, the feature from different source images can be fused together in the internal of CNN in an unsupervised learning fashion. Therefore, compared with feature-level and decision-level fusion methods, our approach is easier to implement and low computational complexity. And to pixel-level fusion, we employ unsupervised learning style to fuse images from different sources instead of utilizing hand-engineered pixel-level fusion algorithms such as discrete wavelet transform (DWT) pixel-level image fusion methodologies.

\subsection{Regions of Interest Proposal}
As you can see in the ROIs proposal module in Figure \ref{pipeline}, given an image, the ROIs proposal algorithms can output a set of class-independent locations that are likely to contain objects. Different from the exhaustive search "sliding window" paradigm which will propose every possible candidate locations and generate around $10^4-10^5$ windows per image, ROIs proposal methods try to reach high object recall with considerably fewer windows. In the most popular object detectors such as R-CNN \cite{GirshickRichSegmentation} and fast R-CNN \cite{Girshick2016}, they select Selective Search \cite{Uijlings2013SelectiveRecognition} method as their ROIs proposal module.

The Selective Search \cite{Uijlings2013SelectiveRecognition} is a ROIs proposal that combines the intuitions of bottom-up segmentation and exhaustive search. The whole algorithm can be simplified as follows. Firstly, \cite{Felzenszwalb2004EfficientSegmentation} algorithm is adopted to create initial regions. Then the similarities between all neighbor regions are calculated and the two most similar regions are grouped together. After that, the new similarities are calculated between the resulting region and its neighbors. In this iterative manner, the process of grouping the most similar regions is repeated until the whole image becomes a single region. Finally, the object location boxes can be extracted from each region. Because of this hierarchical grouping process, the generated locations come from all scales.

\subsection{Network Architecture}
The great success of Convolutional Neural Networks (CNNs) in recent years aroused broader interest in CNNs-based generic object detection among researchers. In typically, a CNN comprises a stack of convolutional and pooling layers. The convolutional layer can generate feature maps by convolving the input feature maps or image with a set of learnable kernels. And the pooling layer can pool information of a given region on output feature maps in order to achieve down sampling and expansion of the receptive field.	

The most typical CNNs-based object detector is the R-CNN \cite{GirshickRichSegmentation}, which utilize Selective Search method to generate a set of ROIs proposal from input image and then feed each ROI to a CNN to obtain final results. However, this paradigm is slow, because lots of heavily overlapped ROIs have to go through the CNN separately and thus a large amount of redundant computation is consumed. SPP-net \cite{He2015} and fast R-CNN \cite{Girshick2016} successful solved this problem by proposing an Spatial Pyramid Pooling (SPP) and ROI pooling, respectively. They suggested the whole image can go through CNN once and the final decision is made at the last feature maps produced by the CNN by using their proposed pooling strategies.

Our proposed framework is illustrated in Figure \ref{pipeline}. The fused three channel image is firstly fed into the CNN feature extractor to generate conv feature maps. It should be noted that the final conv feature maps in our project are also the fusing results of the three types of images by unsupervised learning. After which, for each ROIs generated by the ROIs proposal, we conduct an ROI pooling process directly on the conv feature maps instead of an input image to extract a fixed length feature vector. The reason to choose ROI pooling instead of SPP is that the gradients can propagate to the CNN layers in training stage and this can help CNN learn how to fuse the multiple channel-independent images in an unsupervised fashion. Finally, the extracted vector need to be sent to a fully connected neural network which has two output ports where one is for classification and another one is for bounding boxes regression. 

Taking the trade-off between accuracy and computational complexity into account, the VGGM from \cite{Chatfield2014} is selected as the CNN feature extractor in our framework. Specifically, The VGGM is a shallow version of VGG16 \cite{Simonyan2015VERYRECOGNITION} and wider version of AlexNet\cite{Krizhevsky2012ImageNetNetworks}, but faster than VGG16 as well as more accurate than AlexNet. More detail about the VGGM configuration can be seen in Table \ref{network_configuration}.
\begin{table*}[tp]
\caption{Network configuration: The complete network architecture contains two modules, first module is called \textit{CNN feature extractor} which includes 5 convolutional layers (conv 1-5), second module is the \textit{ROI classification and regression} which has an ROI pooling layer and 4 fully connected layers. }
\centering
\resizebox{\textwidth}{!}{
\begin{tabular}{|c|c|c|c|c|c|c|c|c|c|c|c|c|c|c|}
\hline
\textbf{Name}& Conv1 & Norm1 & Pool1& Conv2& Norm2& Pool2& Conv3& Conv4 & Conv5& ROI Pooling & FC6 &FC7& Cls& Bbox \\ \hline \hline
\textbf{Input Channels}& 3 & 96 & 96& 96& 256& 256& 256& 512 & 512& 512 & 36 &4096& 1024& 1024 \\ \hline 
\textbf{Output Channels}& 96 & 96& 96& 256& 256& 256& 512 & 512& 512 & 36 &4096 & 1024& 2& 8  \\ \hline 
\textbf{Kernel Size}& $7\times7$ &  & $3\times3$  & $5\times5$ &  & $3\times3$ & $3\times3 $ & $3\times3$ & $3\times3$  & $6\times6 $  & &  & &  \\ \hline 
\textbf{Type}& conv& LRN &  max-pool& conv& LRN &  max-pool& conv & conv&  conv & ROI-pool & fc & fc & fc & fc \\ \hline 
\textbf{Stride} & 2 &  & 2& 2&  & 2& 1& 1 & 1&  &  &  &  & \\ \hline 
\textbf{Pad} &  &  & & 1&  & & 1& 1 & 1&  &  &  &  &\\ \hline 
\textbf{Activation function}& relu &  & & relu&  & &relu& relu & relu&  &  & &  & \\ \hline 
\textbf{Dropout}&  &  & & &  & & &  & &  & $\checkmark$ & $\checkmark$ &  &\\ \hline 
\end{tabular}
}

\label{network_configuration}
\end{table*}

\subsection{Training Details}
\subsubsection{Transfer Learning}
Transferring general information between different data source for related tasks is an effective techniques to help deal with insufficient training data and overfitting in deep learning scenario. For instance, In the variants of R-CNN, they firstly train the model on the large ImageNet \cite{RussakovskyImageNetChallenge} dataset and then finetune it on their domain-specific dataset. However, they were only limited to transferring information between RGB(visible image) models.

The target training dataset includes the visible images, IR images, and generated motion maps, whose data distribution is significantly different to the large-scale public visible image datasets, such as ImageNet \cite{RussakovskyImageNetChallenge}. Our goal is to leverage the CNN model gain essential common knowledge from a large-scale visible dataset and then transfer these information for accelerating training in the domain-specific dataset as well as boosting overall performance.

Based on the general transfer learning techniques, the VGGM model is pre-trained on the large-scale RGB image dataset, ImageNet, which contains most common objects and scenes in daily life. Before training the network on the fused dataset, the weights of conv1 to conv5 in the network are initialized by transferring learned weights. Unlike some prior work, we do not freeze the lower layers of CNN and allow our network to adapt the new data distribution in an end-to-end learning setting.
\subsubsection{Loss Function}
As shown in Table \ref{network_configuration}, the network has two output heads. The first is for classifying each ROI, which will output a discrete probability distribution over two categories( background and target). And the second is for regressing the bounding box offsets of ROI where for each category, it will output a tuple of $(t_x,t_y,t_w,t_h)$, the elements indicate the shift value relative to the central coordinate, height and width of original proposal ROI. 

Same to \cite{Girshick2016}, for classification, we use the negative log likelihood objective:
\begin{equation}
L_{cls}(p,u) = -log (p_u)
\end{equation}
where, $p$ represents the predicted probability of one of categories and $u$ is the ground truth class.

For regression, the smooth $L_1$ loss function is used:
\begin{equation}
L_{bbox}(t^u,v) = \sum_{i\in \{x,y,w,h \}} smooth_{L_1}(t^u_i-v_i)
\end{equation}
in which $t^u$ is the bounding box offsets of the $u$ class. And $v$ is the true offsets.

In training stage, both of them will be jointed together as follow:
\begin{equation}
L(p,u,t^u,v) = L_{cls}(p,u) + \lambda[u=1]L_{bbox}(t^u,v)
\end{equation}
$u=1$ means only when the class is target, the bounding box regression can be trained. The $ \lambda$ is used to control balance between classification and regression, we set it as 1 in all experiments.
\section{Experiments}
\subsection{Dataset}
\begin{figure}[tph]
\includegraphics[width=0.8\linewidth]{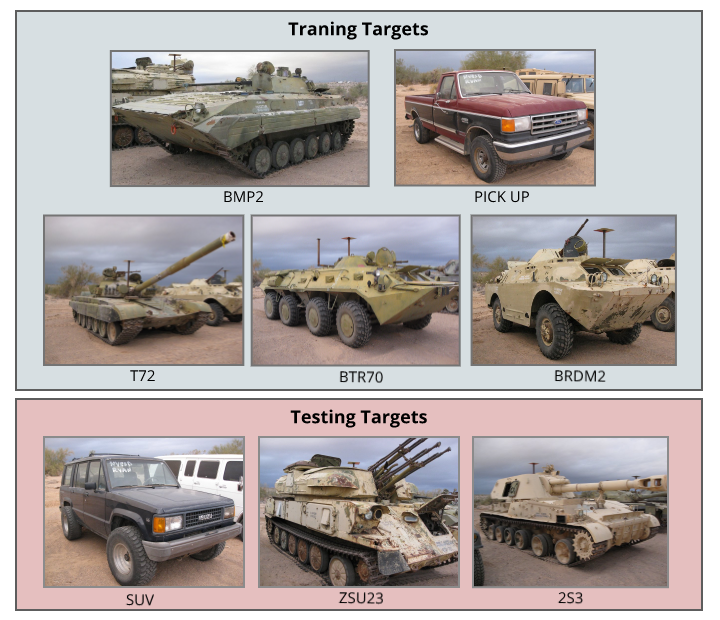}
\centering
\caption{Appearance of targets in training dataset and testing dataset.}
\label{train_test}
\end{figure}
 We evaluate our approach on a public released ATR database from Military Sensing Information Analysis Center (SENSIAC). This database package contains 207 GB of MWIR imagery and 106 GB of visible imagery along with ground truth data. All imagery was taken using commercial cameras operating in the MWIR and visible bands. The types of targets are various, which include people, foreign military vehicles, and civilian vehicles. The datasets were collected during both daytime and night and the distance between cameras and targets varied from 500 to 5000 meters.

In our experiments, we only consider the objects of vehicle, and split 5 types of vehicles as training targets and 3 types of vehicles as testing targets, their name and appearance are showed in Figure \ref{train_test}. And we select each type of vehicles with 3 different range of distance between cameras and targets (1000, 1500 and 2000 meters).  It should be noted that no matter how many fine-grained types of vehicle it has, we treated them as one class, "vehicle". Thus, the problem become a binary (vehicle and background) object detection problem. Moreover, because the format of raw data is video, we sampled the images at every 5 frames to maximize the difference between each frame. In total, we used 4573 images as training data and 2812 images as testing data.

\subsection{Experimental Setup}
Our framework is implemented by using Caffe deep learning toolbox \cite{JiaCaffe:}. For the training machine, we use a normal computer with a NVIDIA GeForce GTX 1080 GPU, a Intel Core i7 CPU and 32 GB Memory.
For the setup of hyper-parameters, we also follow the fast R-CNN  \cite{Girshick2016}, where we train all the networks each for 40000 iterations with initial learning rate 0.001 and 0.0001 for the last 10000 iterations, momentum 0.9 and weight decay 0.0005.
\subsection{Evaluation}
\subsubsection{Metrics}
For all the metrics, we considered the detection as true or false positives based on whether the area of overlap with ground truth bounding boxes exceed $0.5$. The overlap area can be calculated by the below equation:
\begin{equation}
a_o = \frac{area(B_p \cap B_{gt})}{area(B_p \cup B_{gt})}
\end{equation}
Where $a_o$ denotes the overlap area, $B_p$ and $B_{gt}$ denote the predicted bounding box and ground truth bounding box, respectively.

\textbf{Mean Average Precision(mAP)} is a golden standard metric for evaluating the performance of an object detection algorithm, where it first calculates the average precision (AP) of each class and then average all the obtained AP values. Because there is only one class (vehicle) in our experiments, we select AP as one of the evaluation metrics. The AP value can be easily obtained by computing the area under the precision-recall curve.

\textbf{Top1 Precision} is a metric that is widely used in classification tasks, where the probability of multiple classes is predicted and one having the highest score is selected, then the top1 precision score is computed as the numbers a predicted label matched the target label, divided by the number of whole data. In our case, there is only one target in each image. Thus, we can employ top1 precision metric in experiments to evaluate the performance of our framework in a practical scenario.
\subsubsection{Results and Analysis}
\begin{figure}[thp]	
\centering
\includegraphics[width=0.9\linewidth]{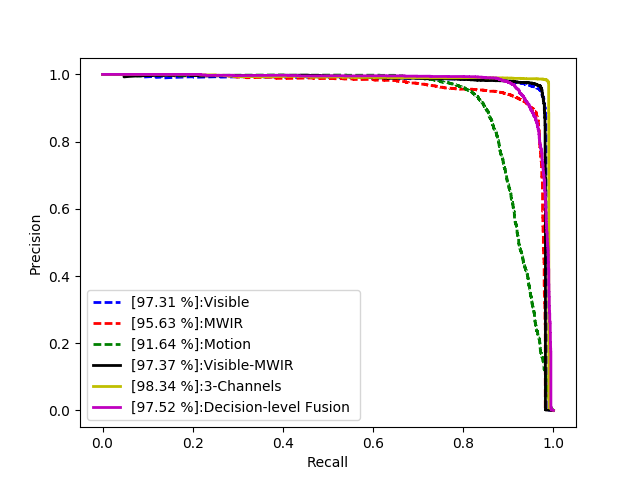}
\caption{Average precision (AP) comparison between different experimental designs. Independent input of visible, MWIR and Motion image, fusion image of visible and MWIR image (Visible-MWIR), fusion image of visible, MWIR and Motion (3-Channels) and decision-level fusion, respectively.}
\label{ap}
\end{figure}

We perform six incremental experiment designs to examine the effectiveness of our fusion method. And at the beginning, we attempt to see the performance of the detection algorithms on three type of images (Visible, MWIR and Motion) independently. Because all of the independent images are single-channel format and the input format requirement of CNN is three-channels image, we generate the desired images by duplicating the single-channel image in three times. After that, we fuse visible and MWIR images together and examine whether the fused image without short-temporal information can boost the overall performance or not. To meet the requirements of CNN, we duplicate visible channel for twice. Then, we incorporate the short-temporal information, Motion image, and generate our complete fused image, 3-channels image. In addition, we also test the decision-level fusion method, where we combine the three outputs of three single networks on three type of independent images.

\newcommand{\ra}[1]{\renewcommand{\arraystretch}{#1}}

\begin{table}[thp]
\caption{Performance comparison on accuracy and time cost of different methods.}
\centering
\ra{1.3}
\resizebox{\linewidth}{!}{
\begin{tabular}{@{}crrcrrrc@{}}\toprule
\textbf{Methods}& \multicolumn{2}{c}{\textbf{Accuracy (\%)}} & \phantom{abc}& \multicolumn{3}{c}{\textbf{Running Time (s/image)}} &
\phantom{abc} \\
\cmidrule{2-3} \cmidrule{5-7} 
& AP & Top1 && ROIs Proposal & Networks & Overall \\ \midrule
Visible & 97.31 & 98.04 &&  1.378& 0.164 &1.542 \\
MWIR & 95.63 & 96.91 &&  \textbf{1.144}& 0.069 &1.213 \\
Motion & 91.64 & 92.39 &&  1.167& \textbf{0.038} &\textbf{1.205} \\
Visible-MWIR & 97.37 &98.18 && 1.505 & 0.248 &1.753 \\
3-Channels & \textbf{98.34}& \textbf{98.90} &&  1.272&0.235 &1.507 \\
Decision-level Fusion & 97.52 &97.93 && 3.690 &0.271 &3.961 \\
\bottomrule
\end{tabular}
}

\label{performance}
\end{table}
Figure \ref{ap} shows the AP curves of the six incremental experiments.In independent image experiments, we can see that the CNN-based detector performed well enough in overall, especially for the single visible image which achieved $97.31\%$ average precision and $98.04\%$ top1 accuracy, as shown in accuracy column of Table  \ref{performance}. The visible-MWIR fused image get a better result than the best performance of single image. It should be noted that our 3-channels image achieve both the highest average precision($98.34\%$) and top1 accuracy ($98.90\%$) which means our method only false detect 16 frames in the totally 2812 testing frames. It is also interesting that even thought the average precision of the decision fusion method is higher than the best single image method, but when it comes to practical application, its top1 accuracy is lower than the single visible image approach and it is extremely time-consuming in running time ($3.961s$).

\begin{figure*}[ph]
\centering
\includegraphics[width=0.9\linewidth]{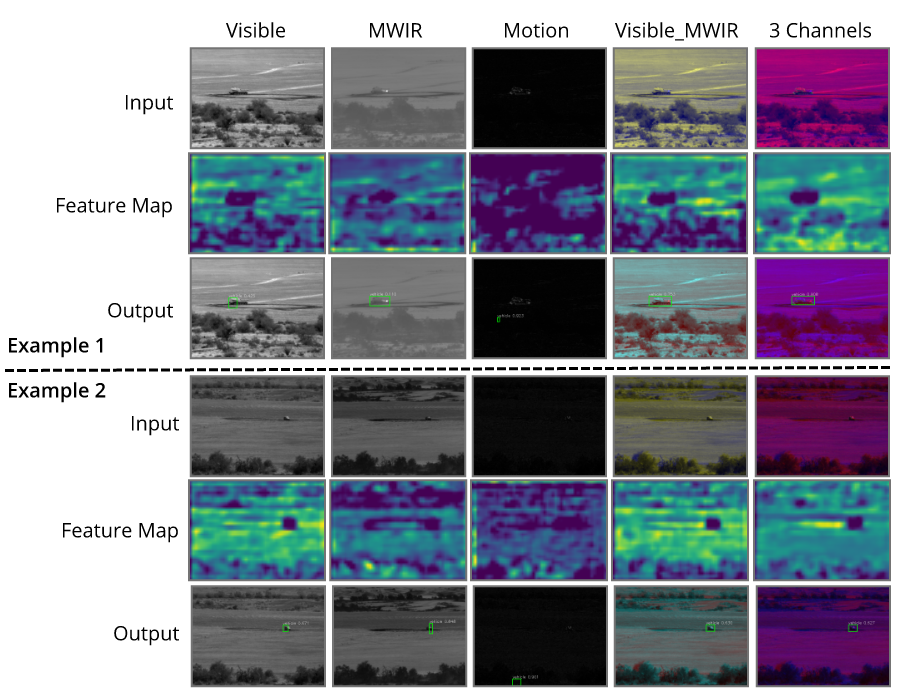}
\caption{Example visualizing results of our framework. Example 1 and 2 demonstrate the performance of different type of inputs in system on large and small object detection, respectively. Different columns denote different types of input image. The raw input image, generated feature map and the final output are showed in consecutive rows. In the final output image, the green bounding box represent the position of object predicted by system. }
\label{visualiztion}
\end{figure*}

To further verify the effectiveness of our unsupervised image fusion method, we visualize the feature map of the last convolutional layer and the final output of our framework in Figure \ref{visualiztion}. The feature map is the output of CNN feature extractor in Figure \ref{pipeline}, and for the fused image, it is the fused high-level features. It could be reasoned that if the object in feature map is segmented clearly, the framework will get a better result. In the examples of Figure \ref{visualiztion}, we can see that the 3-channels can well fuse the complementary information from the three independent images and make its feature map get enhancement. And its final output also verify the fact that the enhanced feature map can boost the performance.

\section{Conclusion}

In this work, an unsupervised learning based image fusion method is proposed to integrate the ATD network, which fused visible, MWIR and motion information effectively. We further adopted state-of-the-art generic object detector for the battle field object detection. We also utilized cross-domain transfer learning techniques to deal with the insufficient data by training the model on large-scale visible image dataset firstly and then fine-tuning on the small-scale fused image dataset. The proposed framework was evaluated with the SENSIAC dataset. It achieved $98.34\%$ average precision and $98.90\%$ top1 accuracy. However, the processing time is still too long for real-time applications. This remains a topic for our future work.

\bibliographystyle{IEEEtranTIE}
\bibliography{Mendeley}\ 

\begin{thebibliography}{10}
\providecommand{\url}[1]{#1}
\csname url@samestyle\endcsname
\providecommand{\newblock}{\relax}
\providecommand{\bibinfo}[2]{#2}
\providecommand{\BIBentrySTDinterwordspacing}{\spaceskip=0pt\relax}
\providecommand{\BIBentryALTinterwordstretchfactor}{4}
\providecommand{\BIBentryALTinterwordspacing}{\spaceskip=\fontdimen2\font plus
\BIBentryALTinterwordstretchfactor\fontdimen3\font minus
  \fontdimen4\font\relax}
\providecommand{\BIBforeignlanguage}[2]{{%
\expandafter\ifx\csname l@#1\endcsname\relax
\typeout{** WARNING: IEEEtran.bst: No hyphenation pattern has been}%
\typeout{** loaded for the language `#1'. Using the pattern for}%
\typeout{** the default language instead.}%
\else
\language=\csname l@#1\endcsname
\fi
#2}}
\providecommand{\BIBdecl}{\relax}
\BIBdecl

\bibitem{Ratches2011ReviewTasks}
J.~A. Ratches, ``{Review of current aided/automatic target acquisition
  technology for military target acquisition tasks},'' \emph{Optical
  Engineering}, vol.~50, \href{http://dx.doi.org/10.1117/1.3601879}{DOI
  10.1117/1.3601879}, no.~7, pp. 072\,001--072\,001, 2011.

\bibitem{Sheikh2005BayesianScenes}
Y.~Sheikh and M.~Shah, ``{Bayesian Object Detection in Dynamic Scenes},'' 2005.

\bibitem{Guo2013FastDetection}
J.-m. Guo, S.~Member, C.-h. Hsia, M.~Ieee, Y.-f. Liu, and S.~Member, ``{Fast
  Background Subtraction Based on a Multilayer Codebook Model for Moving Object
  Detection},'' vol.~23, no.~10, pp. 1809--1821, 2013.

\bibitem{Chen2015ProbabilisticSystems}
\BIBentryALTinterwordspacing
B.-h. Chen and S.-c. Huang, ``{Probabilistic neural networks based moving
  vehicles extraction algorithm for intelligent traffic surveillance
  systems},'' \emph{Information Sciences}, vol. 299,
  \href{http://dx.doi.org/10.1016/j.ins.2014.12.033}{DOI
  10.1016/j.ins.2014.12.033}, pp. 283--295, 2015. [Online]. Available:
  \url{http://dx.doi.org/10.1016/j.ins.2014.12.033}
\BIBentrySTDinterwordspacing

\bibitem{Yun2017SceneCamera}
K.~Yun, J.~Lim, and J.~Y. Choi, ``{Scene conditional background update for
  moving object detection in a moving camera},'' \emph{Pattern Recognition
  Letters}, vol.~88, \href{http://dx.doi.org/10.1016/j.patrec.2017.01.017}{DOI
  10.1016/j.patrec.2017.01.017}, pp. 57--63, 2017.

\bibitem{Hu2015MovingCamera}
\BIBentryALTinterwordspacing
W.-C. Hu, C.-H. Chen, T.-Y. Chen, D.-Y. Huang, and Z.-C. Wu, ``{Moving object
  detection and tracking from video captured by moving camera},''
  \href{http://dx.doi.org/10.1016/j.jvcir.2015.03.003}{DOI
  10.1016/j.jvcir.2015.03.003}, 2015. [Online]. Available:
  \url{http://ac.els-cdn.com/S104732031500053X/1-s2.0-S104732031500053X-main.pdf?_tid=58e2ce5c-249b-11e7-a3f7-00000aab0f6c&acdnat=1492563708_eaac5d28180332af8f627dbfe1c446ba}
\BIBentrySTDinterwordspacing

\bibitem{Francisco2015ForegroundApproximation}
\BIBentryALTinterwordspacing
â.~Francisco, J.~L{\'{o}}pez-Rubio, and E.~L{\'{o}}pez-Rubio, ``{Foreground
  detection for moving cameras with stochastic approximation},'' \emph{Pattern
  Recognition Letters}, vol.~68,
  \href{http://dx.doi.org/10.1016/j.patrec.2015.09.007}{DOI
  10.1016/j.patrec.2015.09.007}, pp. 161--168, 2015. [Online]. Available:
  \url{http://ac.els-cdn.com/S0167865515003177/1-s2.0-S0167865515003177-main.pdf?_tid=249b8706-249b-11e7-abfc-00000aacb362&acdnat=1492563621_7d6892646917decce5a8df7f2bb255ca}
\BIBentrySTDinterwordspacing

\bibitem{Keypoints2004DistinctiveFrom}
S.-i. Keypoints and D.~G. Lowe, ``{Distinctive Image Features from},''
  \emph{International Journal of Computer Vision}, vol.~60,
  \href{http://dx.doi.org/10.1023/B:VISI.0000029664.99615.94}{DOI
  10.1023/B:VISI.0000029664.99615.94}, no.~2, pp. 91--110, 2004.

\bibitem{Dalal2010ToDetection}
N.~Dalal, B.~Triggs, O.~Gradients, D.~Cordelia, N.~Dalal, and B.~Triggs, ``{To
  cite this version : Histograms of Oriented Gradients for Human Detection},''
  pp. 886--893, 2010.

\bibitem{P.Viola2001RapidFeatures}
M.~J. P.~Viola, ``{Rapid Object Detection Using A Boosted Cascade of Simple
  Features},'' \emph{Proceedings of the 2001 IEEE Computer Society Conference
  on Computer Vision and Pattern Recognition},
  \href{http://dx.doi.org/10.1109/CVPR.2001.990517}{DOI
  10.1109/CVPR.2001.990517}, no.~1, pp. 511--518, 2001.

\bibitem{Suykens1999LeastClassifiers}
J.~A.~K. Suykens and J.~Vandewalle, ``{Least Squares Support Vector Machine
  Classifiers},'' \emph{Neural Processing Letters 9},
  \href{http://dx.doi.org/10.1023/A:1018628609742}{DOI
  10.1023/A:1018628609742}, pp. 293--300, 1999.

\bibitem{Breiman2001Randomforest2001}
L.~Breiman, ``{Randomforest2001},''
  \href{http://dx.doi.org/10.1017/CBO9781107415324.004}{DOI
  10.1017/CBO9781107415324.004}, pp. 1--33, 2001.

\bibitem{Felzenszwalb2009ObjectModels}
\BIBentryALTinterwordspacing
P.~F. Felzenszwalb, R.~B. Girshick, D.~Mcallester, and D.~Ramanan, ``{Object
  Detection with Discriminatively Trained Part Based Models},'' \emph{IEEE
  Transactions on Pattern Analysis and Machine Intelligence}, vol.~32,
  \href{http://dx.doi.org/10.1109/TPAMI.2009.167}{DOI 10.1109/TPAMI.2009.167},
  no.~9, pp. 1--20, 2009. [Online]. Available:
  \url{http://cs.brown.edu/~pff/papers/lsvm-pami.pdf}
\BIBentrySTDinterwordspacing

\bibitem{RussakovskyImageNetChallenge}
\BIBentryALTinterwordspacing
O.~Russakovsky, J.~Deng, H.~Su, J.~Krause, S.~Satheesh, S.~Ma, Z.~Huang,
  A.~Karpathy, A.~Khosla, M.~Bernstein, A.~C. Berg, L.~Fei-Fei, O.~Russakovsky,
  J.~Deng, H.~Su, J.~Krause, S.~Satheesh, S.~Ma, Z.~Huang, A.~Karpathy,
  A.~Khosla, M.~Bernstein, A.~C. Berg, and L.~Fei-Fei, ``{ImageNet Large Scale
  Visual Recognition Challenge}.'' [Online]. Available:
  \url{https://arxiv.org/pdf/1409.0575.pdf}
\BIBentrySTDinterwordspacing

\bibitem{Everingham2010TheChallenge}
M.~Everingham, L.~Van~Gool, C.~K. I~Williams, J.~Winn, A.~Zisserman,
  M.~Everingham, L.~K. Van Gool~Leuven, B.~CKI~Williams, J.~Winn, and
  A.~Zisserman, ``{The PASCAL Visual Object Classes (VOC) Challenge},''
  \emph{Int J Comput Vis}, vol.~88,
  \href{http://dx.doi.org/10.1007/s11263-009-0275-4}{DOI
  10.1007/s11263-009-0275-4}, pp. 303--338, 2010.

\bibitem{LinMicrosoftContext}
\BIBentryALTinterwordspacing
T.-Y. Lin, M.~Maire, S.~Belongie, L.~Bourdev, R.~Girshick, J.~Hays, P.~Perona,
  D.~Ramanan, C.~L. Zitnick, and P.~Dol{\'{i}}, ``{Microsoft COCO: Common
  Objects in Context}.'' [Online]. Available:
  \url{https://arxiv.org/pdf/1405.0312.pdf}
\BIBentrySTDinterwordspacing

\bibitem{Sermanet2014OverFeat:Networks}
P.~Sermanet, D.~Eigen, X.~Zhang, M.~Mathieu, R.~Fergus, and Y.~Lecun,
  ``{OverFeat: Integrated Recognition, Localization and Detection using
  Convolutional Networks},'' 2014.

\bibitem{GirshickRichSegmentation}
\BIBentryALTinterwordspacing
R.~Girshick, J.~Donahue, T.~Darrell, and J.~Malik, ``{Rich feature hierarchies
  for accurate object detection and semantic segmentation}.'' [Online].
  Available: \url{http://www.cs.berkeley.edu}
\BIBentrySTDinterwordspacing

\bibitem{Uijlings2013SelectiveRecognition}
\BIBentryALTinterwordspacing
J.~R.~R. Uijlings, K.~E.~A. Van De~Sande, T.~Gevers, and A.~W.~M. Smeulders,
  ``{Selective Search for Object Recognition},'' \emph{Int J Comput Vis}, vol.
  104, \href{http://dx.doi.org/10.1007/s11263-013-0620-5}{DOI
  10.1007/s11263-013-0620-5}, pp. 154--171, 2013. [Online]. Available:
  \url{http://disi.}
\BIBentrySTDinterwordspacing

\bibitem{Girshick2016}
R.~Girshick, ``{Fast R-CNN},'' in \emph{Proceedings of the IEEE International
  Conference on Computer Vision},
  \href{http://dx.doi.org/10.1109/ICCV.2015.169}{DOI 10.1109/ICCV.2015.169},
  2016.

\bibitem{He2015}
K.~He, X.~Zhang, S.~Ren, and J.~Sun, ``{Spatial Pyramid Pooling in Deep
  Convolutional Networks for Visual Recognition},'' \emph{IEEE Transactions on
  Pattern Analysis and Machine Intelligence}, vol.~37,
  \href{http://dx.doi.org/10.1109/TPAMI.2015.2389824}{DOI
  10.1109/TPAMI.2015.2389824}, no.~9, pp. 1904--1916, 2015.

\bibitem{RenFasterNetworks}
S.~Ren, K.~He, R.~Girshick, and J.~Sun, ``{Faster R-CNN: Towards Real-Time
  Object Detection with Region Proposal Networks}.''

\bibitem{Bhatnagar2015ASurveillance}
G.~Bhatnagar and Z.~Liu, ``{A novel image fusion framework for night-vision
  navigation and surveillance},'' \emph{Signal, Image and Video Processing},
  \href{http://dx.doi.org/10.1007/s11760-014-0740-6}{DOI
  10.1007/s11760-014-0740-6}, 2015.

\bibitem{Niu2012AirborneTransform}
Y.~Niu, S.~Xu, L.~Wu, and W.~Hu, ``{Airborne infrared and visible image fusion
  for target perception based on target region segmentation and discrete
  wavelet transform},'' \emph{Mathematical Problems in Engineering}, vol. 2012,
  \href{http://dx.doi.org/10.1155/2012/275138}{DOI 10.1155/2012/275138}, 2012.

\bibitem{Han2007FusionDetection}
\BIBentryALTinterwordspacing
J.~Han and B.~Bhanu, ``{Fusion of color and infrared video for moving human
  detection},'' \emph{Pattern Recognition}, vol.~40,
  \href{http://dx.doi.org/10.1016/j.patcog.2006.11.010}{DOI
  10.1016/j.patcog.2006.11.010}, pp. 1771--1784, 2007. [Online]. Available:
  \url{https://www.researchgate.net/profile/Bir_Bhanu/publication/222827442_Fusion_of_color_and_infrared_video_for_moving_human_detection/links/0912f50b0b010e4b15000000.pdf}
\BIBentrySTDinterwordspacing

\bibitem{Smeelen2014Semi-hiddenImages}
\BIBentryALTinterwordspacing
M.~A. Smeelen, P.~B.~W. Schwering, A.~Toet, and M.~Loog, ``{Semi-hidden target
  recognition in gated viewer images fused with thermal IR images},''
  \emph{Information Fusion}, vol.~18,
  \href{http://dx.doi.org/10.1016/j.inffus.2013.08.001}{DOI
  10.1016/j.inffus.2013.08.001}, pp. 131--147, 2014. [Online]. Available:
  \url{http://ac.els-cdn.com/S156625351300081X/1-s2.0-S156625351300081X-main.pdf?_tid=ea9a0eba-24a9-11e7-bc7a-00000aab0f26&acdnat=1492569966_af49728e4b8bc2c1eb9379e1c1606efb}
\BIBentrySTDinterwordspacing

\bibitem{Felzenszwalb2004EfficientSegmentation}
\BIBentryALTinterwordspacing
P.~F. Felzenszwalb and D.~P. Huttenlocher, ``{Efficient Graph-Based Image
  Segmentation},'' \emph{International Journal of Computer Vision}, vol.~59,
  no.~2, pp. 167--181, 2004. [Online]. Available:
  \url{http://fcv2011.ulsan.ac.kr/files/announcement/413/IJCV(2004)%20Efficient%20Graph-Based%20Image%20Segmentation.pdf}
\BIBentrySTDinterwordspacing

\bibitem{Chatfield2014}
\BIBentryALTinterwordspacing
K.~Chatfield, K.~Simonyan, A.~Vedaldi, and A.~Zisserman, ``{Return of the Devil
  in the Details: Delving Deep into Convolutional Nets},'' \emph{arXiv preprint
  arXiv: {\ldots}}, \href{http://dx.doi.org/10.5244/C.28.6}{DOI
  10.5244/C.28.6}, pp. 1--11, 2014. [Online]. Available:
  \url{http://arxiv.org/abs/1405.3531}
\BIBentrySTDinterwordspacing

\bibitem{Simonyan2015VERYRECOGNITION}
\BIBentryALTinterwordspacing
K.~Simonyan and A.~Zisserman, ``{VERY DEEP CONVOLUTIONAL NETWORKS FOR
  LARGE-SCALE IMAGE RECOGNITION},'' 2015. [Online]. Available:
  \url{https://arxiv.org/pdf/1409.1556/}
\BIBentrySTDinterwordspacing

\bibitem{Krizhevsky2012ImageNetNetworks}
A.~Krizhevsky, I.~Sutskever, and G.~E. Hinton, ``{ImageNet Classification with
  Deep Convolutional Neural Networks},'' \emph{Advances In Neural Information
  Processing Systems},
  \href{http://dx.doi.org/http://dx.doi.org/10.1016/j.protcy.2014.09.007}{DOI
  http://dx.doi.org/10.1016/j.protcy.2014.09.007}, pp. 1--9, 2012.

\bibitem{JiaCaffe:}
\BIBentryALTinterwordspacing
Y.~Jia, E.~Shelhamer, J.~Donahue, S.~Karayev, J.~Long, R.~Girshick,
  S.~Guadarrama, and T.~Darrell, ``{Caffe: Convolutional Architecture for Fast
  Feature Embedding *}.'' [Online]. Available:
  \url{https://arxiv.org/pdf/1408.5093.pdf}
\BIBentrySTDinterwordspacing

\end{thebibliography}

\end{document}